\documentclass[11pt,a4paper]{article}
\usepackage[hyperref]{acl2020}
\usepackage{microtype,times,latexsym,booktabs,amsmath,multirow,tikz,framed,arydshln,mdwlist,linguex}

\newlength{\catheight}
\newcommand{\ourmodel}{CAt~\raisebox{-0.1ex}{\includegraphics[height=1.2\catheight]{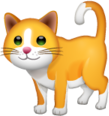}}} 
\definecolor{aspectcolor}{HTML}{fb7e5e}

\DeclareMathOperator{\rbf}{rbf}
\DeclareMathOperator{\att}{att}
\DeclareMathOperator{\softmax}{softmax}
\DeclareMathOperator*{\argmax}{argmax}

\newcommand*{\coloremph}[2]{%
  \tikz[baseline]\node[rectangle, fill=#1, rounded corners, inner sep=2pt, outer sep=0.pt, anchor=base,text height=\catheight, text depth=0.2ex]{#2};%
}

\aclfinalcopy 
\def\aclpaperid{1636} 

\title{Embarrassingly Simple Unsupervised Aspect Extraction}
\author{St\'ephan Tulkens\\
  CLiPS \\
  University of Antwerp \\
  Belgium \\
  \texttt{\normalsize stephan.tulkens@uantwerpen.be} \\\And
  Andreas van Cranenburgh \\
  Department of Information Science \\
  University of Groningen \\
  The Netherlands \\
  \texttt{\normalsize a.w.van.cranenburgh@rug.nl} \\}
\date{}

\begin{document}
\maketitle
\begin{abstract}
We present a simple but effective method for aspect identification
in sentiment analysis.
Our unsupervised method only requires word embeddings and a POS tagger,
and is therefore straightforward to apply to new domains and languages.
We introduce Contrastive Attention (\ourmodel), a novel single-head attention mechanism based on an RBF kernel, which gives a considerable boost in performance and makes the model interpretable.
Previous work relied on syntactic features and complex neural models.
We show that given the simplicity of current benchmark datasets
for aspect extraction, such complex models are not needed.
The code to reproduce the experiments reported in this paper
is available at \url{https://github.com/clips/cat}.
\end{abstract}

\section{Introduction}

We consider the task of unsupervised aspect extraction from text.
In sentiment analysis, an aspect can intuitively be defined as
a dimension on which an entity is evaluated (see \autoref{figaspectexample}).
While aspects can be concrete (e.g., a laptop battery), they can also be subjective (e.g., the loudness of a motorcycle).
Aspect extraction is an important subtask
of aspect-based sentiment analysis.
However, most existing systems are supervised~\citep[for an overview, cf.~][]{zhang2018deep}.
As aspects are domain-specific, supervised systems that rely on strictly lexical cues to differentiate between aspects are unlikely to transfer well between different domains~\citep{rietzler2019adapt}.
Another reason to consider the unsupervised extraction of aspect terms is the scarcity of training data for many domains (e.g., books), and, more importantly, the complete lack of training data for many languages.
Unsupervised aspect extraction has previously been attempted with topic models~\citep{mukherjee2012aspect}, topic model hybrids~\citep{garcia2018w2vlda}, and restricted Boltzmann machines~\citep{wang2015sentiment}, among others.
Recently, autoencoders using attention mechanisms~\citep{he2017unsupervised,luo2019unsupervised} have also been proposed as a method for aspect extraction, and have reached state of the art performance on a variety of datasets.
These models are unsupervised in the sense that they do not require labeled data, although they do rely on unlabeled data to learn relevant patterns.
In addition, these are complex neural models with a large number of parameters. We show that a much simpler model suffices for this task.

\begin{figure}[t]
    \begin{framed}
    The two things that really drew me to \emph{vinyl} \\
    were the \coloremph{aspectcolor}{expense} and the \coloremph{aspectcolor}{inconvenience}.
    \end{framed}
    \caption{An example of a sentence expressing two aspects (red) on a target (italics). Source: 
    \url{https://www.newyorker.com/cartoon/a19180}}
    \label{figaspectexample}
\end{figure}

\begin{figure}[t]\centering
    \begin{tikzpicture}[
        vec/.style={
    		rectangle,
    		rounded corners,
    		draw=black, 
    		very thick,
    		text centered,
    		text width=40mm,
    		minimum height=7mm,
        },
        box/.style={ minimum height=15mm, text width=25mm, },
    	red/.style={ fill=red!10 },
    	blue/.style={ fill=blue!10 },
    	green/.style={ fill=green!10 },
    	yellow/.style={ fill=yellow!10 },
        ]

    \path (4, 4) node[vec, box, yellow] (A) {A: aspect vectors};
    \path (0, 4) node[vec, box, blue] (S) {S: sentence (word vectors)};
    \path (2, 1.75) node[vec, red] (att) {att: attention vector};
    \path (2, 0.5) node[vec, green] (d) {d: sentence summary};
    \path (0.5, -0.5) node (label1) {food};
    \path (2, -0.5) node (label2) {staff};
    \path (3.5, -0.5) node (label3) {ambience};

    \draw [->, very thick] (S) -> (A);

    \path (2, 3) node[circle, draw=black, font=\scriptsize] (rbf) {RBF};
    
    \draw [very thick] (A) -- (rbf);
    \draw [very thick] (S) -- (rbf);
    \draw [->, very thick] (rbf) -> (att.north);
    
    \draw [->, very thick] (att) -> (d);
    
    \draw [->, dotted, thick] (d) -> (label1);
    \draw [->, dotted, thick] (d) -> (label2);
    \draw [->, dotted, thick] (d) -> (label3);
\end{tikzpicture}
    \caption{An overview of our aspect extraction model.}\label{figmodel}
\end{figure}
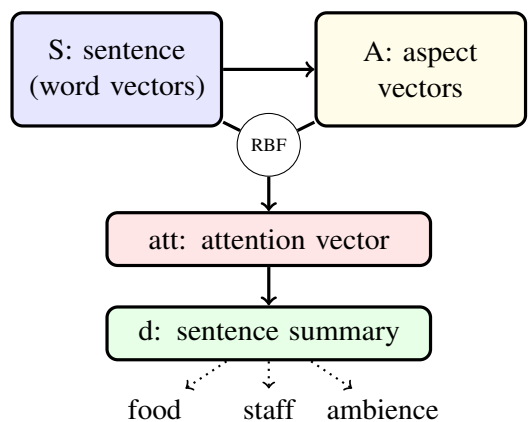

We present a simple unsupervised method for aspect extraction which only requires a POS tagger and in-domain word embeddings, trained on a small set of documents.
We introduce a novel single-head attention mechanism, Contrastive Attention (\ourmodel), based on Radial Basis Function (RBF) kernels.
Compared to conventional attention mechanisms~\citep{weston2014memory, sukhbaatar2015end},
\ourmodel\ captures more relevant information from a sentence.
Our method outperforms more complex methods,
e.g., attention-based neural networks~\citep{he2017unsupervised, luo2019unsupervised}.
In addition, our method automatically assigns aspect labels,
while in previous work, labels are manually assigned to aspect clusters.
Finally, we present an analysis of the limitations of our model, and propose some directions for future research.

\section{Method}

Like previous methods~\citep{hu2004mining, xu2013mining},
our method (see \autoref{figmodel}) consists of two steps:
extraction of candidate aspect terms and assigning aspect labels to instances.
Both steps assume a set of in-domain word embeddings, which we train using \texttt{word2vec}~\citep{mikolov2013efficient}. 
We use a small set of in-domain documents, containing about 4 million tokens for the restaurant domain.

\paragraph{Step 1: aspect term extraction}

In previous work~\citep{hu2004mining, xu2013mining}, the main assumption has been that nouns that are frequently modified by sentiment-bearing adjectives (e.g., good, bad, ugly) are likely to be aspect nouns.
We experimented with this notion and devised a labeling strategy in which aspects are extracted based on their co-occurrence with seed adjectives.
However, during experimentation we found that for the datasets in this paper, the most frequent nouns were already good aspects;
any further constraint led to far worse performance on the development set.
This means that our method only needs a POS tagger to recognize nouns, not a full-fledged parser.
Throughout this paper, we use \texttt{spaCy}~\citep{spacy2} for tokenization and POS tagging.
In \autoref{secanalysis},  we investigate how these choices impact performance.

\begin{figure}[t]
    \input{fig/attention}
    \caption{Examples of Contrastive Attention ($\gamma$=$.03$)}
    \label{figattention}
\end{figure}

\paragraph{Step 2: aspect selection using Contrastive Attention}

We use a simple of form of attention, similar to the attention mechanism used in memory networks~\citep{weston2014memory, sukhbaatar2015end}.
With an attention mechanism, a sequence of words, e.g., a sentence or a document, is embedded into a matrix $S$, which is operated on with an aspect $a$ to produce a probability distribution, $\att$.
Schematically:

\begin{equation}
    \att = \softmax(aS)
\end{equation}

$\att$ is then multiplied with $S$ to produce an informative summary with respect to the aspect $a$:

\begin{equation}
    d = \sum_{i} \att_i S_i
\end{equation}

Where $d$ is the weighted sentence summary.
There is no reason to restrict $a$ to be a single vector: when replaced by a matrix of queries, $A$, the equation above gives a separate attention distribution for each aspect, which can then be used to create different summaries, thereby keeping track of different pieces of information.
In our specific case, however, we are interested in tracking which words elicit aspects, regardless of the aspect to which they belong.
We address this by introducing Contrastive Attention~(\ourmodel), a way of calculating attention that integrates a set of query vectors into a single attention distribution.
It uses an RBF kernel, which is defined as follows:

\begin{equation}
    \rbf(x, y, \gamma) = \exp(-\gamma ||x - y||^{2}_{2})
\end{equation}

where, $x$ and $y$ are vectors, and $\gamma$ is a scaling factor, which we treat as a hyperparameter.
An important aspect of the RBF kernel is that it turns an arbitrary unbounded distance, the squared euclidean distance in this case, into a bounded similarity. 
For example, regardless of $\gamma$, if $x$ and $y$ have a distance of 0, their RBF response will be 1. 
As their distance increases, their similarity decreases, and will eventually asymptote towards 0, depending on $\gamma$.
Given the RBF kernel, a matrix $S$, and a set of aspect vectors $A$, attention is calculated as follows:

\begin{equation}
\att = \frac{\sum_{a \in A} \rbf(w, a, \gamma)}{\sum_{w \in S} \sum_{a \in A} \rbf(w, a, \gamma)}
\end{equation}

The attention for a given word is thus the sum of the RBF responses of all vectors in $A$, divided by the sum of the RBF responses of the vectors to all vectors in $S$.
This defines a probability distribution over words in the sentence or document, where words that are, on average, more similar to aspects, get assigned a higher score.

\paragraph{Step 3: assigning aspect labels}

After reweighing the word vectors, we label each document based on the cosine similarity between the weighted document vector $d$
and the label vector.

\begin{equation}\label{eq:labeling}
    \hat y = \argmax_{c \in C}(\cos(d, \vec{c}))
\end{equation}

Where $C$ is the set of labels, i.e., \{\textsc{food}, \textsc{ambience}, \textsc{staff}\}.
In the current work, we use word embeddings of the labels as the targets.
This avoids the inherent subjectivity
of manually assigning aspect labels,
the strategy employed in previous work~\citep{he2017unsupervised,luo2019unsupervised}.

\begin{table}[t]\centering
\begin{tabular}{lrr}  & \textbf{Train} & \textbf{Test}  \\ \midrule
Citysearch (\citeyear{ganu2009beyond})  &       & 1,490 \\
SemEval (\citeyear{pontiki2014semeval}) & 3,041 &   402 \\
SemEval (\citeyear{pontiki2015semeval}) & 1,315 &   250 \\
\end{tabular}
\caption{The number of sentences in each of the datasets after removing sentences that did not express exactly one aspect in our set of aspects.}\label{tab:corpora}
\end{table}

\begin{table}[t]\centering
\begin{tabular}{lrrr}
\textbf{Method} & \textbf{P} & \textbf{R} & \textbf{F} \\ \midrule
SERBM (\citeyear{wang2015sentiment})    & 86.0  & 74.6  & 79.5  \\
ABAE (\citeyear{he2017unsupervised})    & \textbf{89.4}  & 73.0  & 79.6  \\
W2VLDA (\citeyear{garcia2018w2vlda})    & 80.8  & 70.0  & 75.8  \\
AE-CSA (\citeyear{luo2019unsupervised}) & 85.6  & 86.0  & 85.8 \\
Mean      & 78.9  & 76.9  & 77.2  \\
Attention & 80.5  & 80.7  & 80.6  \\
\ourmodel & 86.5  & \textbf{86.4}  & \textbf{86.4}
\end{tabular}
\caption{Weighted macro averages across all aspects on the test set of the Citysearch dataset.}\label{tblresultsmacro}
\end{table}

\section{Datasets}
We use several English datasets of restaurant reviews for the aspect extraction task.
All datasets have been annotated with one or more sentence-level labels,
indicating the aspect expressed in that sentence
(e.g., the sentence ``The sushi was great'' would be assigned the label \textsc{food}).
We evaluate our approach on the Citysearch dataset \cite{ganu2009beyond}, which uses the same labels as the SemEval datasets.
To avoid optimizing for a single corpus, we use the restaurant subsets of the SemEval 2014~\citep{pontiki2014semeval} and SemEval 2015~\citep{pontiki2015semeval} datasets as development data.
Note that, even though our method is completely unsupervised, we explicitly allocate test data to ensure proper methodological soundness, and do not optimize any models on the test set.
Following previous work~\citep{he2017unsupervised, ganu2009beyond},
we restrict ourselves to sentences that only express exactly one aspect;
sentences that express more than one aspect, or no aspect at all,
are discarded.
Additionally, we restrict ourselves to three labels:
\textsc{food}, \textsc{service}, and \textsc{ambience}.
We adopt these restrictions in order to compare to other systems.
Additionally, previous work~\citep{brody2010unsupervised} reported that the other labels, \textsc{anecdotes} and \textsc{price}, were not reliably annotated.
\autoref{tab:corpora} shows statistics of the datasets.

\begin{table}[t]\centering
\begin{tabular}{lrrr}
\textbf{Method} & \textbf{P} & \textbf{R} & \textbf{F} \\ \midrule
\multicolumn{4}{c}{Aspect: \textsc{food}} \\
SERBM (\citeyear{wang2015sentiment}) & 89.1  & 85.4  & 87.2  \\
ABAE (\citeyear{he2017unsupervised})      & 95.3  & 74.1  & 82.8  \\
W2VLDA (\citeyear{garcia2018w2vlda})    & \textbf{96.0} & 69.0  & 81.0  \\
AE-CSA (\citeyear{luo2019unsupervised})   & 90.3  & \textbf{92.6}  & 91.4  \\
Mean      & 92.4  & 73.5  & 85.6  \\
Attention & 86.7  & 89.5  & 88.1  \\
\ourmodel & 91.8  & 92.4 & \textbf{92.1}  \\ \midrule
\multicolumn{4}{c}{Aspect: \textsc{staff}} \\
SERBM (\citeyear{wang2015sentiment}) & 81.9  & 58.2  & 68.0  \\
ABAE (\citeyear{he2017unsupervised})     & 80.2  & 72.8  & 75.7  \\
W2VLDA (\citeyear{garcia2018w2vlda})   & 61.0  & \textbf{86.0}  & 71.0  \\
AE-CSA (\citeyear{luo2019unsupervised})   & 92.6  & 75.6  & 77.3  \\
Mean      & 55.8  & 85.7  & 67.5  \\
Attention & 74.4  & 69.3  & 71.8  \\
\ourmodel & \textbf{82.4}  & 75.6  & \textbf{78.8}  \\ \midrule
\multicolumn{4}{c}{Aspect: \textsc{ambience}}\\
SERBM (\citeyear{wang2015sentiment} & 80.5  & 59.2  & 68.2  \\
ABAE (\citeyear{he2017unsupervised})      & 81.5  & 69.8  & 74.0  \\
W2VLDA (\citeyear{garcia2018w2vlda})   & 55.0  & 75.0  & 64.0  \\
AE-CSA (\citeyear{luo2019unsupervised})   & \textbf{91.4}  & 77.9  & \textbf{77.0}  \\
Mean      & 58.7  & 56.1  & 57.4  \\
Attention & 67.1  & 65.7  & 66.4  \\
\ourmodel & 76.6  & \textbf{80.1}  & 76.6
\end{tabular}
\caption{Precision, recall, and F-scores 
on the test set of the Citysearch dataset.}\label{tblresults}
\end{table}

\section{Evaluation}

We optimize all our models on SemEval '14 and '15 training data;
the scores on the Citysearch dataset do not reflect any form of optimization with regards to performance.
We optimize the hyperparameters of each model separately
(i.e., the number of aspect terms and $\gamma$ of the RBF kernel), leading to the following hyperparameters:
For the regular attention, we select the top 980 nouns as aspect candidates.
For the RBF attention, we use the top 200 nouns and a $\gamma$ of .03.

We compare our system to four other systems.
W2VLDA~\citep{garcia2018w2vlda} is a topic modeling approach that biases word-aspect associations by computing the similarity from a word to a set of aspect terms.
SERBM~\citep{wang2015sentiment} a restricted Boltzmann Machine (RBM) that learns topic distributions, and assigns individual words to these distributions. 
In doing so, it learns to assign words to aspects.
We also compare our system to two attention-based systems. First, ABAE~\citep{he2017unsupervised}, which is an auto-encoder that learns an attention distribution over words in the sentence by simultaneously considering the global context and aspect vectors.
In doing so, ABAE learns an attention distribution, as well as appropriate aspect vectors.
Second, AE-CSA~\cite{luo2019unsupervised}, which is a hierarchical model which is similar to ABAE. 
In addition to word vectors and aspect vectors, this model also considers sense and sememe~\cite{bloomfield1926set} vectors in computing the attention distribution.
Note that all these systems, although being unsupervised, do require training data, and need to be fit to a specific domain.
Hence, all these systems rely on the existence of in-domain training data on which to learn reconstructions and/or topic distributions.
Furthermore, much like our approach, ABAE, AE-CSA, and W2VLDA rely on the availability of pre-trained word embeddings.
Additionally, AE-CSA needs a dictionary of senses and sememes, which might not be available for all languages or domains.
Compared to other systems, our system does require a UD POS tagger to extract frequent nouns.
However, this can be an off-the-shelf POS tagger, since it does not need to be trained on domain-specific data.

We also compare our system to a baseline based on the mean of word embeddings, a version of our system using regular attention, and a version of our system using Contrastive Attention (\ourmodel).
The results are shown in \autoref{tblresults}.
Because of class imbalance
(60~\% of instances are labeled \textsc{food}),
the F-scores from \ref{tblresults} do not give a representative
picture of model performance.
Therefore, we also report weighted macro-averaged scores in \autoref{tblresultsmacro}.

Our system outperforms ABAE, AE-CSA, and the other systems, both in weighted macro-average F1 score, and on the individual aspects.
In addition, \ref{tblresultsmacro} shows that the difference between ABAE and SERBM is smaller than one would expect based on the F1 scores on the labels, on which ABAE outperforms SERBM on \textsc{Staff} and \textsc{Ambience}.
The Mean model still performs well on this dataset,
while it does not use any attention or knowledge of aspects.
This implies that aspect knowledge is probably not required to perform well on this dataset; focusing on lexical semantics is enough.

\begin{figure}[t]\centering
    \includegraphics[width=\linewidth]{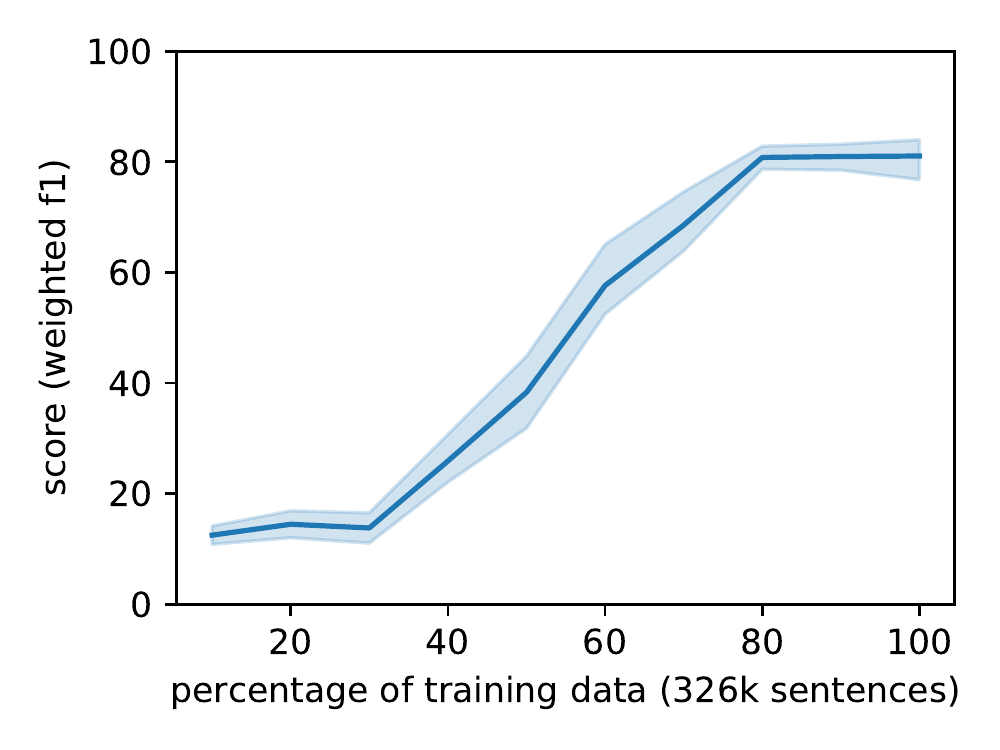}
    \caption{A learning curve on the restaurant data, averaged over 5 embedding models.}
    \label{figlearningcurve}
\end{figure}

\section{Analysis}\label{secanalysis}
We perform an ablation study to see the influence of each component of our system;
specifically, we look at the effect of POS tagging, in-domain word embeddings, and the amount of data on performance.

Only selecting the most frequent words as aspects, regardless of their POS tag, had a detrimental effect on performance, giving an F-score of 64.5 ($\Delta$-21.9), while selecting nouns based on adjective-noun co-occurrence had a smaller detrimental effect, giving an F-score of 84.4 ($\Delta$-2.2), higher than ABAE and SERBM.

Replacing the in-domain word embeddings trained on the training set
with pretrained GloVe embeddings~\citep{pennington2014glove}\footnote{%
    Specifically, the \texttt{glove.6B.200D} vectors from \url{https://nlp.stanford.edu/projects/glove/}}
had a large detrimental effect on performance, dropping the F-score to 54.4 ($\Delta$-32); this shows that in-domain data is important.

To investigate how much in-domain data is required to achieve good performance, we perform a learning curve experiment (\autoref{figlearningcurve}).
We increase the training data in 10\% increments,
training five \texttt{word2vec} models at each increment.
As the figure shows, only a modest amount of data
(about 260k sentences) is needed to tackle this specific dataset.

\begin{table}[t]
\begin{tabular}{ll}
\textbf{Phenomenon}  & \textbf{Example}         \\
\midrule
OOV         & ``I like the Somosas''  \\
Data Sparsity & ``great Dhal'' \\
Homonymy    & ``Of course''      \\
Verb $>$ Noun & ``Waited for food'' \\
Discourse   & ``She didn't offer dessert''  \\
Implicature & ``No free drink''
\end{tabular}
\caption{A categorization of observed error types.}
\label{taberrors}
\end{table}

To further investigate the limits of our model, we perform a simple error analysis on our best performing model.
\autoref{taberrors} shows a manual categorization
of error types.
Several of the errors relate to Out-of-Vocabulary (OOV) or low frequency items, such as the words `Somosas' (OOV) and `Dhal' (low frequency).
Since our model is purely based on lexical similarity, homonyms and polysemous words can lead to errors. 
An example of this is the word `course,'
which our model interprets as being about food.
As the aspect terms we use are restricted to nouns,
the model also misses aspects expressed in verbs, such as ``waited for food.''
Finally, discourse context and implicatures often lead to errors.
The model does not capture enough context or world knowledge to infer that `no free drink' does not express an opinion about drinks, but about service.

Given these errors, we surmise that our model will perform less well in domains in which aspects are expressed in a less overt way.
For example, consider the following sentence from a book review~\citep{kirkus2019review}:

\ex. As usual, Beaton conceals any number of surprises behind her trademark wry humor.

This sentence touches on a range of aspects, including writing style, plot, and a general opinion on the book that is being reviewed.
Such domains might also require the use of more sophisticated aspect term extraction methods.

However, it is not the case that our model necessarily overlooks implicit aspects. For example, the word ``cheap''  often signals an opinion about the price of something. 
As the embedding of the word ``cheap'' is highly similar to that of ``price'' our model will attend to ``cheap'' as long as enough price-related terms are in the set of extracted aspect terms of the model.

In the future, we would like to address the limitations of the current method, and apply it to
datasets with other domains and languages.
Such datasets exist, but we have not yet evaluated our system on them due to the lack of sufficient unannotated in-domain data in addition to annotated data.

Given the performance of \ourmodel, especially compared to regular dot-product attention, it would be interesting to see how it performs as a replacement of regular attention in supervised models, e.g., memory networks~\citep{weston2014memory, sukhbaatar2015end}.
Additionally, it would be interesting to see why the attention model outperforms regular dot product attention. 
Currently, our understanding is that the dot-product attention places a high emphasis on words with a higher vector norm; words with a higher norm have, on average, a higher inner product with other vectors.
As the norm of a word embedding directly relates to the frequency of this word in the training corpus, the regular dot-product attention naturally attends to more frequent words.
In a network with trainable parameters, such as ABAE~\citep{he2017unsupervised}, this effect can be mitigated by finetuning the embeddings or other weighting mechanisms.
In our system, no such training is available, which can explain the suitability of \ourmodel\  as an unsupervised aspect extraction mechanism.

\section{Conclusion}

We present a simple model of aspect extraction that uses a frequency threshold for candidate selection together with a novel attention mechanism based on RBF kernels, together with an automated aspect assignment method.
We show that for the task of assigning aspects to sentences in the restaurant domain,
the RBF kernel attention mechanism outperforms a regular attention mechanism, as well as more complex models based on auto-encoders and topic models.

\section*{Acknowledgments}
We are grateful to the three reviewers for their feedback.
The first author was sponsored by a Fonds Wetenschappelijk Onderzoek (FWO) aspirantschap.

\bibliography{ref}
\bibliographystyle{aclnatbib}

\end{document}